\documentclass[5p,times,procedia]{elsarticle}
\usepackage{ecrc}
\volume{00}

\firstpage{1}

\journalname{Procedia Manufacturing}

\runauth{Deshpande et al.}

\jid{promfg}

\usepackage{amssymb}
\usepackage[figuresright]{rotating}
\usepackage{hyperref}
\hypersetup{colorlinks,allcolors=black}      

\usepackage{amsmath}
\usepackage{eqnarray}

\begin{document}
\begin{frontmatter}



\dochead{48th SME North American Manufacturing Research Conference, NAMRC 48, Ohio, USA}%

\title{One-Shot Recognition of Manufacturing Defects in Steel Surfaces}


\author[a]{Aditya M. Deshpande \corref{cor1} \fnref{cor2}}
\author[a]{Ali A. Minai \corref{cor1}}
\author[a]{Manish Kumar \corref{cor1}}
\address[a]{University of Cincinnati\\
            Cincinnati, Ohio 45221, USA}

\cortext[cor1]{Email: deshpaad@mail.uc.edu, \{ali.minai, manish.kumar\}@uc.edu
\newline
Email addresses are given in order of author names.}
\fntext[cor2]{Corresponding author}

\begin{abstract}
Quality control is an essential process in manufacturing to make the product defect-free as well as to meet customer needs. The automation of this process is important to maintain high quality along with the high manufacturing throughput. With recent developments in deep learning and computer vision technologies, it has become possible to detect various features from the images with near-human accuracy. However, many of these approaches are data intensive. Training and deployment of such a system on manufacturing floors may become expensive and time-consuming.
The need for large amounts of training data is one of the limitations of the applicability of these approaches in real-world manufacturing systems. In this work, we propose the application of a Siamese convolutional neural network to do one-shot recognition for such a task. Our results demonstrate how one-shot learning can be used in quality control of steel by identification of defects on the steel surface. This method can significantly reduce the requirements of training data and can also be run in real-time.

\end{abstract}


% \cortext[cor1]{Corresponding author. Tel.: +0-000-000-0000 ; fax: +0-000-000-0000.}

\end{frontmatter}

\section{Introduction}
\label{intro}

\begin{figure*}[htbp]
    \centering
    \includegraphics{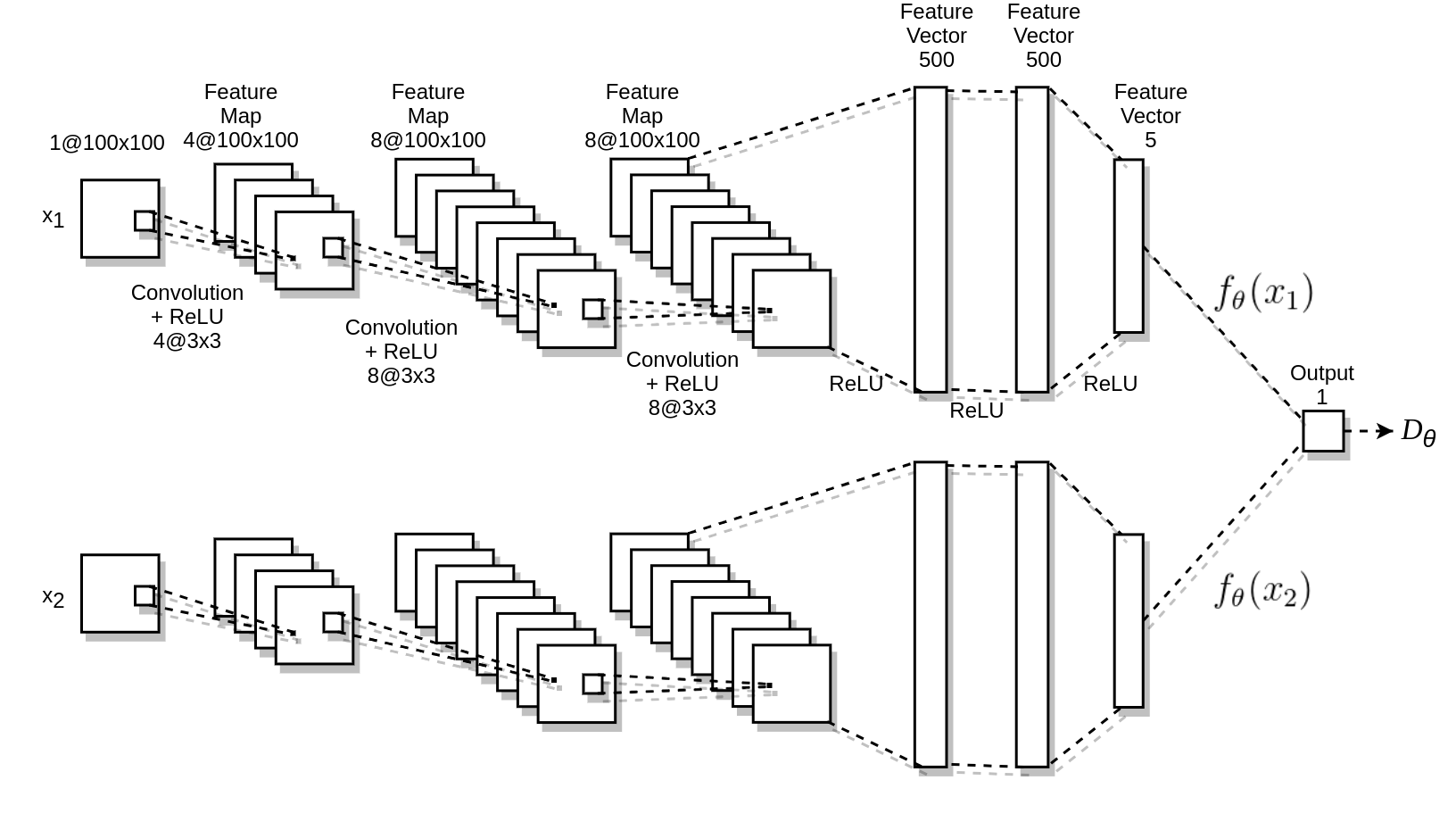}
    \caption{Siamese Network Architecture. $x_1$ and $x_2$ form the pair of input images to the two neural network modules. Output is the euclidean distance between the $5$ dimensional encoded vector of the last layer of each module. Feature map details are provided as `$N_{channel}@W_{channel}\times H_{channel}$'.\label{network_architecture}}
\end{figure*}

To ensure customer satisfaction and reduce manufacturing cost, quality control plays a major role. Often, with a human in the loop, this process consumes a lot of time. With large production requirements and increasing complexity of industrial ecosystems, the human operator's ability to recognize and address the quality has been outpaced. To address this limitation, the automation of quality control is one of the requirements. This automation is done by tracking of the parameters of interest and quantifying their deviations from the desired values. Industry 4.0 and Industrial Internet of Things (IIoT) have resulted in the modernization of manufacturing practices. IIoT has catalyzed instrumentation, monitoring, and analytics in the industry.

IIoT has facilitated the collection of large amounts of data from various sensors and manufacturing processes. This has laid the foundation for the use of data-intensive approaches such as deep learning on the factory floor for monitoring and inspection tasks \cite{pfrommer2018optimisation, meng2018enhancing}. This has paved the way for various innovations in smart manufacturing.

Recently, the field of deep learning and computer vision has produced several pivotal advances that address complex problems. Deep neural networks have solved challenging problems in visual perception \cite{redmon2016you, ren2015faster}, speech-recognition \cite{zhang2018deep}, language understanding \cite{vaswani2017attention, devlin2018bert} and robot autonomy \cite{levine2016end}. These techniques leverage the expressiveness of neural network architectures as powerful and flexible function approximators. Deep neural networks form the basis to learn sophisticated characteristics of the input given large amounts of data. As a result, the field of computer vision is also shifting from statistical methods to deep neural network-based approaches. 

Computer vision methods can be used for non-invasive inspection of the manufacturing output. The quality of prediction of the deep learning-based vision systems is highly dependent on training data. Although IIoT has enabled a large amount of data collection and storage, this data may require annotations. Labeling the data can be expensive and time-consuming. In the case of growing manufacturing facilities, if the manufacturing of a new product is launched, the inspection requirements for such a product may be completely different. Thus, new training data will be required to train a new model. As a result, training and deployment of the deep learning solutions in manufacturing environments can be challenging. These limitations of deep learning-based methods form the motivation of this work. In this paper, we present the novel application of one-shot recognition for steel surface defect detection. The results show the effectiveness of this approach to recognize various defects in steel surfaces by significantly reducing the training data requirements. Even with one sample of a particular class, this approach is able to effectively identify the defect belonging to that class. To the best of our knowledge, this work is first of its kind demonstrating the application of one-shot recognition for quality inspection in steel surfaces.

This paper has been organized in the following order: Section \ref{review} is a brief literature review of the research of artificial intelligence and smart manufacturing. Details of our application of one-shot recognition of surface defects using the Siamese network are presented in section \ref{saimese}. Section \ref{dataset_details} provides the details of the dataset used in this work. Section \ref{results} presents the experimentation details and results. Section \ref{conclusion} gives the conclusion and future work directions.

\begin{figure*}[ht]
    \centering
    \includegraphics[width=\linewidth]{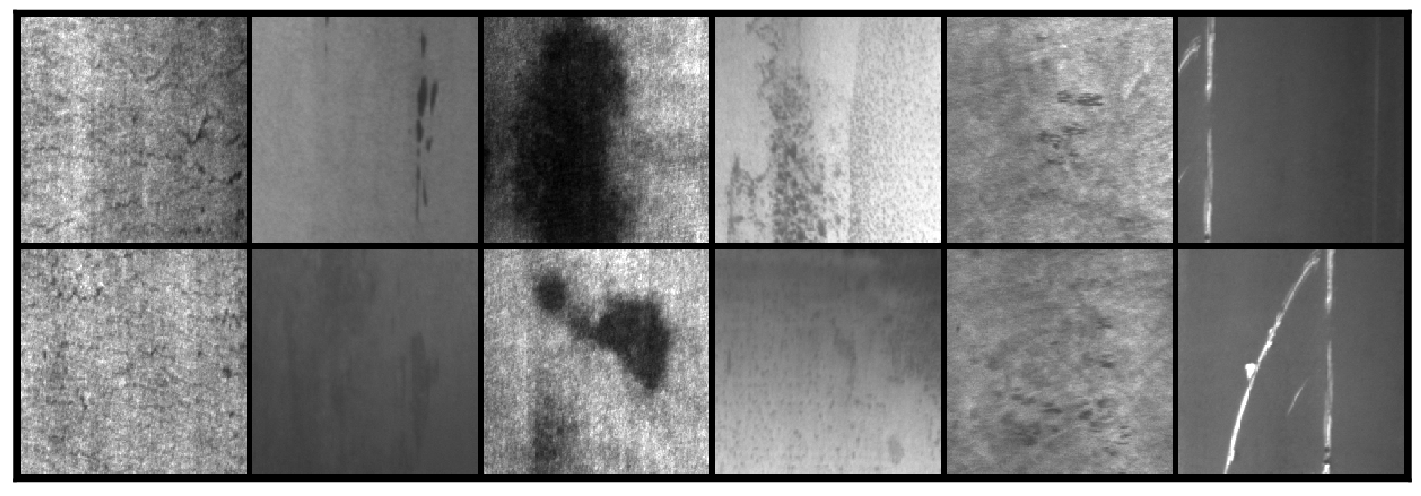}
    \caption{Random samples from the NEU surface defect dataset for each class in columns from left to right: crazing (Cr), inclusion (In), patches (Pa), pitted surface (PS), rolled-in scale (RS) and scratches (Sc) \label{dataset_samples}}
\end{figure*}

\section{Literature review}
\label{review}

There is an unprecedented increase in sensory data as a result of Industry 4.0 and IIoT. Decisions in intelligent manufacturing are influenced by the information collected from all across the manufacturing facilities including manufacturing equipment, manufacturing process, labor activity, product line, and environmental conditions. Machine learning has revolutionized data interpretation approaches. Advancements in deep learning and computer vision have provided robust solutions for difficult problems including object detection \cite{redmon2016you, ren2015faster}, object tracking \cite{choi2018context}, anomaly detection \cite{kiran2018overview} and feature extraction \cite{ronneberger2015u}. Vision-based inspection is classified as nondestructive evaluation technique in manufacturing industry. With combination of recent advancements in computer vision, these inspection processes can be automated and improved without compromising product quality.

To design robust automated non-invasive vision systems for quality control, interdisciplinary knowledge of manufacturing and advanced image-processing techniques is essential. The work in \cite{huang2015automated} presented a detailed overview of inspection tasks that can be potentially automated using vision techniques in the semiconductor industry. An elegant solution for manufacturing defect inspection using convolutional neural networks (CNN) and transfer learning on X-ray images was presented in \cite{ferguson2018detection}. Authors have used Mask Region-based CNN \cite{he2017mask} for this application. This method can perform multiple defect detection as well as segmentation of the same simultaneously in the input image. This application used the GRIMA database of X-ray images (GDXray) for casting and welding \cite{mery2015gdxray} to demonstrate the effectiveness of this approach. Computer vision has been applied for the detection of damage and cracks in concrete surfaces. In one of the early studies applying image processing to detect defects in concrete surface presented a comparative study of various image processing techniques including fast Haar transform, fast Fourier transform, Sobel edge detector, and Canny edge detector \cite{abdel2003analysis}. A robust approach using the deep learning-based crack classification of the concrete surface was presented recently in \cite{cha2017deep}. The authors used a deep CNN and presented a comparative study of their approach with traditional methods including Sobel and Canny edge detection. CNNs were found to be capable of performing without any failures under a wide range of image conditions for crack detection.

Work in \cite{richter2017development} presented novel and integrative intelligent optical inspection approach to monitoring the quality of printed circuit boards (PCBs) in manufacturing lines. The author in this work also emphasized the use of deep neural networks for non-invasive vision-based inspection. Several other machine learning approaches were applied to monitor PCB manufacturing in \cite{vafeiadis2018framework}. This paper presented a detailed comparative study of methods including multi-layer perceptrons, support vector machines (SVMs), radial basis function-based SVMs, decision trees, random forest, naive-Bayes classifier, logistic regression and gradient boosting. Another example of quality control with deep learning in PCB manufacturing can be found in \cite{lin2018capacitor}. To enable real-time inspection and localization of various PCB features in the image, authors of this work have trained the YOLO object detector on the annotated data of PCB images.

Surface inspection is an important part of the quality control process in manufacturing. A lot of work is being done to detect surface flaws using deep learning which aids in quality control. Authors in \cite{park2016machine} have trained the neural network on the surface data of six types including wafer, solid color paint, pearl color paint, fabric, stone and wood. Some of the early work on steel surface defect detection with the application of deep CNNs is available in \cite{soukup2014convolutional} which used photometric stereo images of steel to train the network models.
 
A novel architecture of neural network designed for segmentation and localization of the defect on the metallic surfaces is presented in \cite{tao2018automatic}. In this work, a cascaded autoencoder (CASAE) is used in the first stage to localize and extract the features of the defect from the input image followed by the accurate classification of the defect using a compact CNN in the second stage. In a similar context, the application of the U-Net architecture of neural network \cite{ronneberger2015u} has also proven to be very useful for the saliency detection on surfaces. Authors in \cite{huang2018surface} obtained the state-of-the-art results with the U-Net architecture for the detection of defects on magnetic tile surfaces.

Although deep learning has shown great promises for smart manufacturing, it comes with the cost of large data requirements. Since the annotation of the data collected from the manufacturing lines may not always be possible, there is a limitation on the immediate deployment of these systems. To address these issues, there has been recent interest in the research community to develop neural networks that can effectively learn the mapping from sensor space to the target space from small datasets. Transfer learning in deep neural networks is one such step in that direction \cite{yosinski2014transferable, oquab2014learning}. The key idea here is that hidden layers of CNN are generic extractors of the latent features from the data. The transfer learning enables the reuse of a pre-trained neural network after fine-tuning with a relatively small dataset for a new task. Thus, the Imagenet CNN architecture \cite{krizhevsky2012imagenet} which contains more than 60 million parameters may not require training from scratch but only a few thousand training images may be used to learn new classification task.

The approaches like few-shot learning and zero-shot learning can further reduce the data requirements for deep learning tasks \cite{snell2017prototypical, ravi2017few, vinyals2016matching, romera2015embarrassingly}. The few-shot learning uses only a few examples for each category from a dataset (typically less than 10) to learn image classification. Zero-shot learning is designed to capture the knowledge of various attributes in the data during training and use this knowledge in the inference phase to categorize instances among a new set of classes. The one-shot recognition approach of using Siamese neural network architecture is also an excellent example that requires only one data sample \cite{koch2015siamese}. This network has found applications in areas where data available to train the neural networks may be limited. This one-shot recognition approach has proven to be useful in tasks like drug discovery \cite{altae2017low},  natural language processing \cite{neculoiu2016learning}, audio recognition \cite{droghini2019audio} and image segmentation \cite{shaban2017one}. The low training data requirements of this approach make it suitable for visual inspection tasks. In this work, we explore the application of Siamese network-based one-shot recognition for the visual inspection task in smart manufacturing. We show the effectiveness of this method on steel surface defect recognition. The results also include the comparison of this approach with conventional CNN and a simple one-shot learning algorithm of the Nearest-Neighbor algorithm with a single neighbor.

\section{One-shot recognition using Siamese Networks}
\label{saimese}

The key idea behind one-shot image recognition is that given a single sample of the image of a particular class, the network should be able to recognize if the candidate examples belong to the same class or not. The network learns to identify the differences in features of the input image pair in training. During the inference phase, the learned network can be reused with only one example image of a certain class to recognize if the candidate data belongs to the same class or not.

The Siamese network architecture used in this work is shown in fig. \ref{network_architecture}. This model is trained to learn a good representation of defects in steel surfaces. We use the contrastive loss function explained in section \ref{loss_func} for training the network. The model once trained should be able to recognize multiple defects given a single example of each defect.

In fig. \ref{network_architecture}, the two modules of network are identical and share the same weights. Each module can be viewed as a parametric function of weights $\theta$ given by $f_{\theta}: \mathbb{R}^{N} \rightarrow \mathbb{R}^{n}$ and $N>>n$. High dimensional input (image) $\mathbb{R}^{N}$ is reduced to output which is an encoded vector of lower dimension $n$. In this case, $N=100\times 100$ and $n=5$.
The readers should note that the outputs from the two modules from layers with size $n=5$ are referred to as the encoded vectors $f_{\theta}(x_1)$ and $f_{\theta}(x_2)$. The final output of the architecture is the euclidean distance between these encoded vectors. The input to the model is a single channel or grayscale image pairs $x_1$ and $x_2$. Each module being identical has three convolutional layers with a number of feature maps as $4$, $8$ and $8$ from left to right respectively of size $100 \times 100$ each. The convolutional layers are followed by three fully connected layers of size $500$, $500$ and $5$ respectively. The kernel size of $3 \times 3$ is used for convolutions with a stride of $1$. The $ReLU$ activation function is used on the output feature maps from each layer.

\subsection{Contrastive Loss}
\label{loss_func}

For training, we used contrastive loss function \cite{hadsell2006dimensionality, chopra2005learning}. Equation \eqref{contrastive_loss_eq_1} describes the loss function $L(\cdot)$. The loss function is parameterized by the weights of the neural network $\theta$ and the training sample $i$. The $i^{th}$ training sample from the dataset is a tuple $(x_1, x_2, y)^i$ where $x_1$ and $x_2$ are pair of images and the label $y$ is equal to $1$ if $x_1$ and $x_2$ belong to same class and $0$ otherwise.

\begin{equation}
\label{contrastive_loss_eq_1}
    L(\theta, (x_1, x_2, y)^{i}) = y \frac{1}{2} D_{\theta, i}^2 \\
                                + (1-y) \frac{1}{2} (max\{0, m-D_{\theta, i}\})^2
\end{equation}

The first term of the right hand side (RHS) of equation \eqref{contrastive_loss_eq_1} imposes cost on the network if the input image pair $x_1$ and $x_2$ belongs to same class, i.e., $y=1$. The second term penalizes the input sample if the data belongs to different classes $y=0$. $m > 0$ is a margin and its value is constant. The term $D_{\theta, i}$ is explained in equation \eqref{contrastive_loss_eq_2}.

\begin{equation}
\label{contrastive_loss_eq_2}
    D_{\theta, i} = || f_{\theta}(x_1) - f_{\theta}(x_2) ||_{2, i}
\end{equation}

The equation \eqref{contrastive_loss_eq_2} is the Euclidean distance between the $n$ dimensional outputs of neural network modules for the input image pair of $x_1$ and $x_2$ in sample $i$ of the dataset.

For the $i^{th}$ sample with $y=1$, the second term in equation \eqref{contrastive_loss_eq_1} is evaluated to zero. Therefore, the loss value in this case is directly proportional to the square of distance between $f_{\theta}(x_1)$ and $f_{\theta}(x_2)$. The objective is the minimization of the loss, the network weights are learned so as to reduce the distance between the encoded vectors of input samples $x_1$ and $x_2$. Intuitively, this can be understood as the model learning that the two input images are similar. On the other hand, if the input sample has the label of $y=0$, the first term on the RHS is nullified. If $y=0$ and $D_{\theta, i} > m$, the model is not penalized. The penalty is applied only if the Euclidean distance between $f_{\theta}(x_1)$ and $f_{\theta}(x_2)$ is less than the set margin $m$. The objective in this case is to push the encoded vectors $f_{\theta}(x_1)$ and $f_{\theta}(x_2)$ away from each other in the $n$ dimensional space and make the distance between them greater than $m$. One can think of the second term in loss function as the model learning to understand the differences between $x_1$ and $x_2$ which belong to different classes. As a result of this loss function, the Siamese network not only learns to estimate the similarity score of the input pair of images but the loss values of the dissimilar pairs from non-zero second term avoid the collapse of the model to a constant function. For a detailed mathematical explanation of contrastive loss, authors request the readers to refer the paper \cite{hadsell2006dimensionality}.

\section{Dataset}
\label{dataset_details}
We trained our model of the Siamese network using Northeastern University (NEU) surface defect database\footnotemark \cite{song2013noise, he2019end, he2019semi}. This database consists of six classes of surface defects on hot-rolled steel strip, viz., rolled-in scale (RS), patches (Pa), crazing (Cr), pitted surface (PS), inclusion (In) and scratches (Sc). Dataset has 1,800 grayscale images in total with 300 samples each of the six classes. The resolution of each sample image is $200 \times 200$ pixels. Few sample images from the dataset for each class are shown in fig. \ref{dataset_samples}. The dataset images have a variation in illuminations which introduces further challenges for the image recognition task. This variability results in large differences in samples belonging to the same class. Another challenge that can be observed is due to the similarity in images belonging to the different classes as can be seen in fig. \ref{dataset_samples}. For example, the similarity in images belonging to the categories of crazing and rolled-in scale steel surfaces is easily noticeable.

\footnotetext{Data source: \url{http://faculty.neu.edu.cn/yunhyan/NEU_surface_defect_database.html}, last accessed on November 30, 2019.}

\subsection*{Data Augmentation}
To overcome the problem of limited quantity and limited diversity of data, we augment the existing steel surface defect dataset with affine transformations. Each image in the dataset is rotated randomly about its center. The angle of the rotation is chosen uniformly from the set of angles $\{0, \frac{\pi}{2}, \pi, \frac{3\pi}{2}\}$ (in radian). To further augment the data, we also introduce the horizontal and vertical flips in the data, each with a $0.5$ probability. To favor invariance of the model to light conditions, we introduce perturbations in the image brightness. Random value $\beta$ chosen from a uniform distribution in the range of [-10, 10] is added to the image for this perturbation. The equation \eqref{perturb_op} defines the perturbation operation.

\begin{eqnarray}
\beta & \sim & U(-10, 10) \label{random_sampling_op}\\
I_{out} &=& max(min(I_{in} + \beta, 255), 0)  \label{perturb_op}
\end{eqnarray}

In above equations, $I_{in}$ represents input image, $I_{out}$ is the output image. $U(-10, 10)$ represents uniform distribution to sample the scalar value $\beta$.

\section{Experiments and Results}
\label{results}

\begin{figure}
    \centering
    \includegraphics[width=\linewidth]{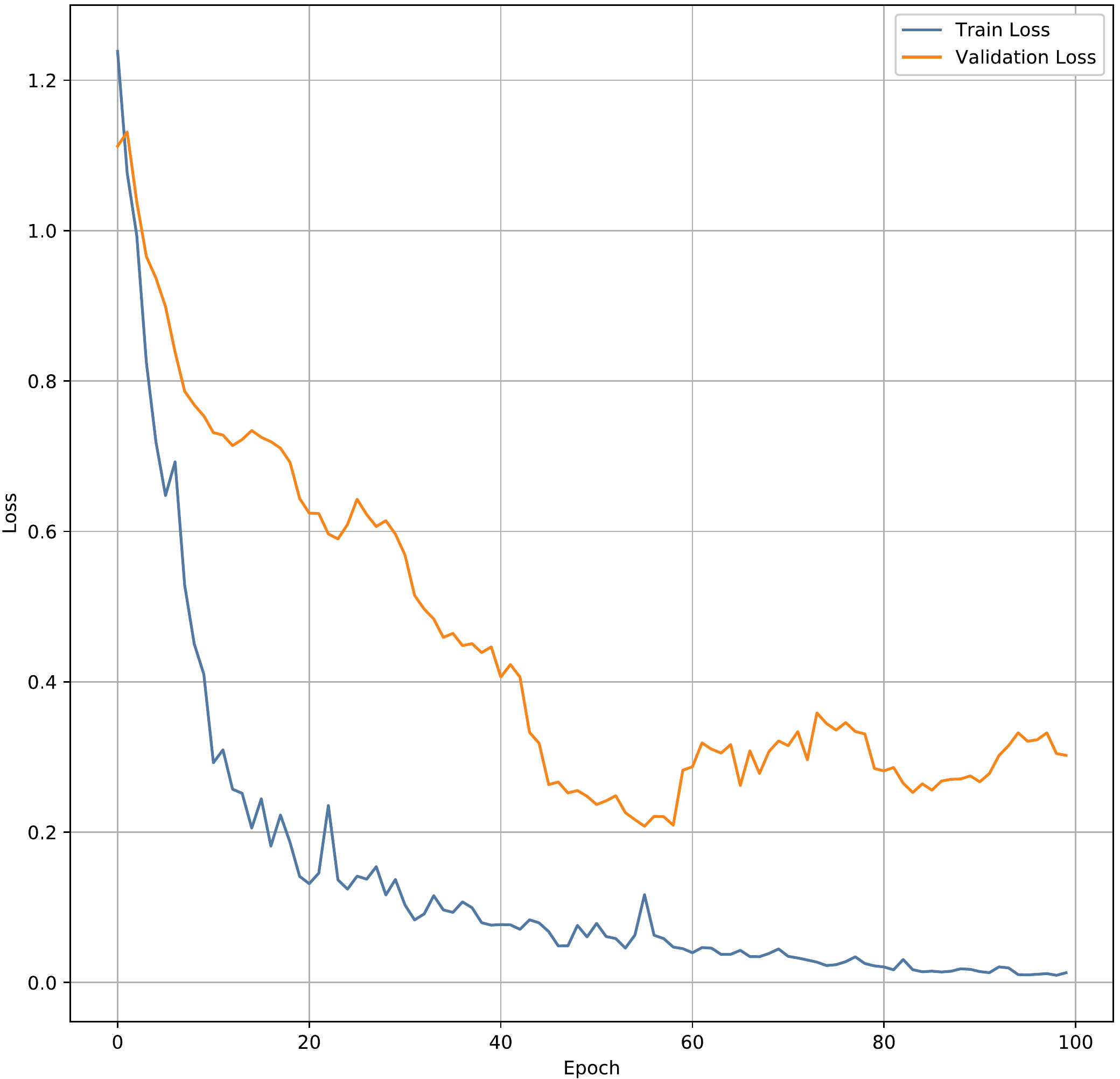}
    \caption{Training and validation curve \label{training_curve}}
\end{figure}

\subsection{Details of the experiment}
The neural network model was trained using the NEU surface defect dataset. The hyperparameter values used for training the network are provided in the table \ref{parameter_table}.

\begin{table}[h!]
    \centering
    \caption{Hyperparameters used to train the Siamese convolutional neural network \label{parameter_table}}
    \begin{tabular}{||r l||} 
         \hline
         Parameter & Value \\ [0.5ex] 
         \hline\hline
         Batch Size & 32 \\
         Number of epochs & 100 \\
         Learning Rate & 5e-4 \\
         Margin $m$ in contrastive loss & 2 \\
         Neural network optimizer & Adam \cite{kingma2014adam} \\
         Adam parameters $(\beta_1, \beta_2)$ & $(0.9, 0.999)$ \\
         \hline
    \end{tabular}
\end{table}

NEU dataset was divided into two sets for one-shot recognition. The training set consisted of the three classes, viz., rolled-in scale, patches, inclusion. The remaining classes of crazing, pitted-surface and scratches were shown to the network in the testing phase for one-shot recognition.

Data samples were chosen randomly during training. While sampling an image pair, the two images were chosen from the same category with a probability of $0.5$ with a corresponding label of $y=1$. Similarly, the images were chosen from two different categories with the remaining probability of $0.5$ with label $y=0$. This tuple of image pair and label $(x_1, x_2, y)$ is then augmented with the transformations described in section \ref{dataset_details}. Before passing the image in the network, the pixel values of each image were normalized to fall in the range of $[-1, 1]$.

The experiments were performed on the Intel-i7 platform with 16GB RAM and NVIDIA RTX 2070. The training with the surface defect dataset was fairly quick. It took approximately $2$ hour for training this architecture from scratch.

\subsection{Results and Discussion}

\begin{figure*}
    \centering
    \includegraphics[width=0.80\textwidth, height=0.60\textwidth]{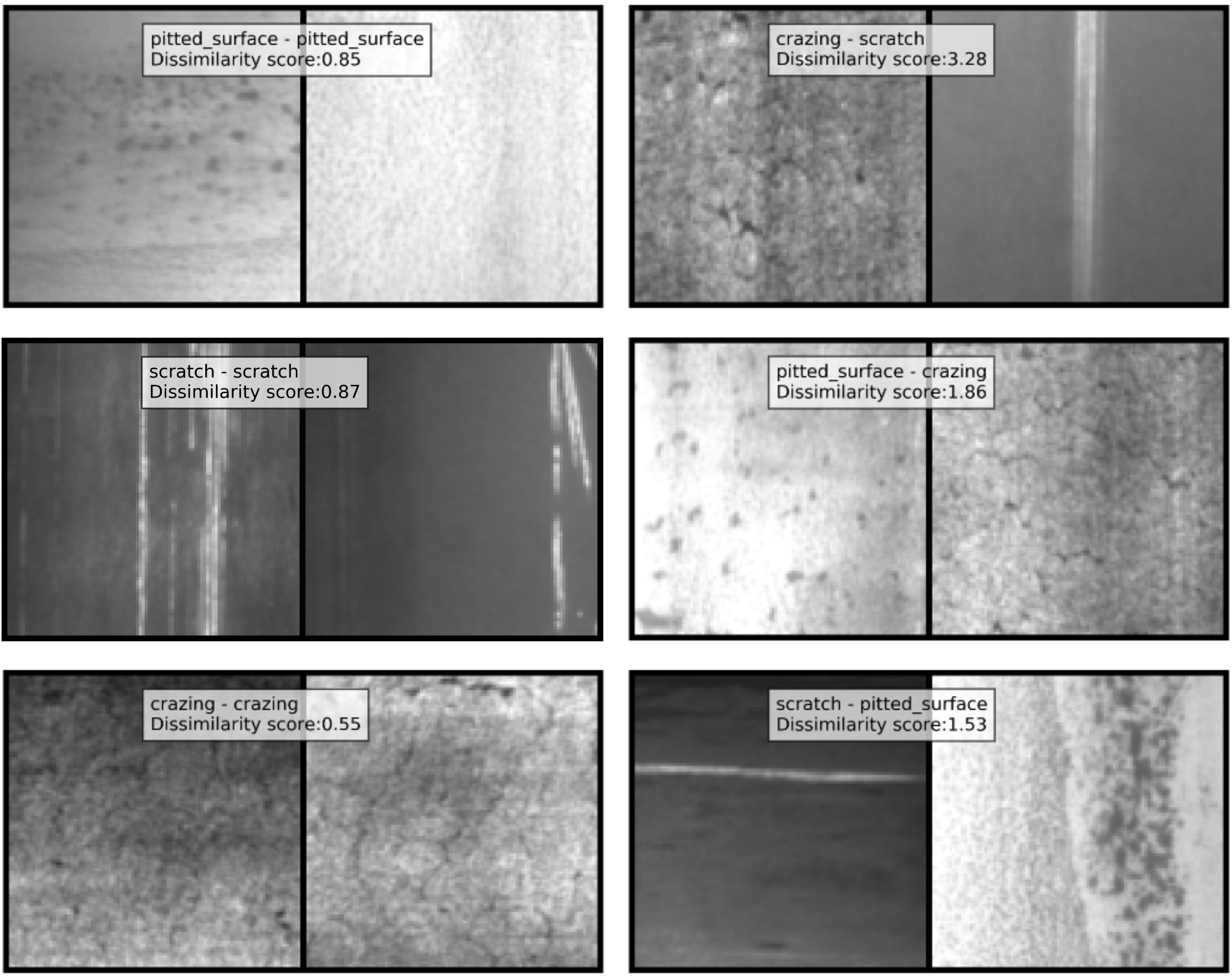}
    \caption{Illustration of results obtained by the network in training phase \label{output_sample_results}}
\end{figure*}

The training and validation curves for the optimization of our model trained using the dataset of $900$ samples of images augmented with transformations as described section \ref{dataset_details} are shown in fig. \ref{training_curve}. The training was done for $100$ epochs with a batch size of $32$. Here, we can see the decreasing trend followed by validation loss along with the training loss as the number of epochs increases. During this learning, the model appears to realize the visual saliencies of the reference image and the candidate image. Thus, loss accumulated during training decreases.

During the testing phase, the images were chosen randomly. These images belong to a different set of classes that were never shown to the network during training. Our trained network model was able to perform the recognition of images in real-time during inference. Each sample took approximately $0.0112$ seconds for evaluation on CPU. The candidate images were classified to be of the same class as the true image used in one-shot recognition based on the value of equation \eqref{contrastive_loss_eq_2} for the image pairs. The margin $m$ was used as the threshold for this decision.

Figure \ref{output_sample_results} illustrates some of the results of the Siamese network evaluated during the testing. Results are presented with class names along with images as well as the dissimilarity score of the image pair. The dissimilarity score is the value of the equation \eqref{contrastive_loss_eq_2} for true image ($x_1$) as well as candidate image ($x_2$).
It can be observed from this figure that images belonging to separate categories have a larger value of the dissimilarity score as compared to images that belong to the same category. The images from dissimilar classes have a score larger than the value of the margin $m$ used in the contrastive loss function. It can be inferred from this observation that the neural network architecture is able to effectively to understand similarities and differences between the features of the input samples.

We compared the results of one-shot recognition with the K-nearest neighbor (KNN) classification algorithm and feed-forward convolutional neural network architecture. The KNN algorithm was chosen since this can form a basic one-shot learning system. With a value of $K=1$, the algorithm was used for image classification of defective surfaces. KNN was shown a single instance of images from each class of the dataset and its test accuracy was evaluated by the proportion of correctly classified test instances. The raw images were used as an input and the euclidean distance between the images was used as a metric in this algorithm for classification of the candidate images into a particular category. We used the KNN implementation from scikit-learn for this purpose \cite{scikitlearn}.

We also compared our approach with a feed-forward CNN classifier \cite{krizhevsky2012imagenet}. The CNN we used for this comparison had a similar architecture as one of the modules from the Siamese network. The input to the network is a single-channel image. The outputs of this network are the class probabilities of the input image belonging to one of the six classes from the surface defect dataset. The network had three convolutional layers with feature maps of $4$, $8$ and $8$ respectively. The size of each feature map was $100 \times 100$. The kernel size of $3 \time 3$ was used for convolutions with the stride of $1$ in these layers. The third convolutional layer was followed by two fully connected layers of size $500$ each. The output layer had 6 neurons. The activation function of ReLU was used except for the output layer which was a sigmoid activation function to represent the class probabilities of the input image. The training set consisted of $80 \%$ of the dataset and remaining data was used for validation and testing for this network. The categorical cross-entropy loss was used for training the CNN along with Adam optimizer. This network was trained for $120$ epochs with a batch size of $128$.

The table \ref{comparison_table} summarizes our testing results for each method on the steel surface defect dataset.

\begin{table}[h!]
    \centering
    \caption{Summary of test results for all the algorithms used in this work \label{comparison_table}}
    \begin{tabular}{||r l||} 
         \hline
         Algorithm & Testing Accuracy (\%) \\ [0.5ex] 
         \hline\hline
         K-nearest neighbor with $k=1$ & $28.22$ \\
         Siamese neural network & $83.22$ \\
         CNN & $93.24$ \\
         \hline
    \end{tabular}
\end{table}

Referring to table \ref{comparison_table}, it can be seen that the KNN algorithm does not work well and shows poor performance in the inference phase. It is clearly not possible to use it in real-world scenarios since it is not optimized for good feature representation of the data as well as the euclidean distance metric is not the appropriate function to quantify the match between of high dimensional image data \cite{lee1991handwritten}. Although the CNN had superior performance, one should also note that one-shot recognition was shown only a single image sample of a new image category to get the observed performance as opposed to $80\%$ of data from each category used to train the CNN.

To have a fair comparison between the proposed Siamese network architecture and the CNN, we also present the result of training both the models with the identical data from NEU dataset in table \ref{fair_comparison_table}. The training set for both the models consisted of $80 \%$ of the NEU dataset from all its six classes and remaining data were used in validation and testing. This table consists of the test accuracy of both models. All the training hyperparameters and loss functions were kept the same as described before in this section for respective neural network models. From the results in table \ref{fair_comparison_table}, it was observed that the CNN and the Siamese network had a competitive performance when trained on the identical data from NEU surface defect dataset. The results in this table also suggest that the Siamese network will converge to the CNN performance as there is increase in the size of dataset used for its training.

\begin{table}[h!]
    \centering
    \caption{Comparison of test results for CNN and Siamese Network \label{fair_comparison_table}}
    \begin{tabular}{||r l||} 
         \hline
         Algorithm & Testing Accuracy (\%) \\ [0.5ex] 
         \hline\hline
         Siamese neural network & $92.55$ \\
         CNN & $93.24$ \\
         \hline
    \end{tabular}
\end{table}

Based on the results observed from the Siamese network for one-shot recognition, it can be said that this approach has the potential for easy and fast deployment on actual factory floors in case of limited training data. With ever-growing production demands and increasing requirements of automation in quality control, this can form a suitable application for the situations where data annotation is difficult or data availability is limited.

\section{Conclusion and Future Work}
\label{conclusion}

In this work, we show the application of one-shot recognition of the Siamese convolutional neural network on steel surfaces. This vision-based approach has two-fold contributions in the automation of quality control. One being non-invasive, the surface quality after production can be remotely inspected without any damage to the steel. The second contribution is the minimal requirement of labeled data for training the images of a new class which makes it easy to adapt this approach for different tasks.

This novel application of deep learning and computer vision paves the way for the development of various new innovations in the manufacturing space. The architecture used for the network presented in this work is not optimum. One of the future directions can be to find out better values of hyperparameters for the dataset. In this case, only single-channel image data of surface defects was used to inspect steel surfaces. The more feature-rich sensor data can be a next good step to explore. One of the apparent directions for future work can be exploring a wider class of texture inspection. The other direction of future work is transferring the learned weights of a pre-trained model like VGG net \cite{Simonyan15} or ResNet \cite{he2016deep} in the framework of one-shot recognition fine-tuned for the domain of vision-based inspection. Apart from images, one-shot recognition can also be used for the identification of similarities or differences in time series data. In this case, applications such as health monitoring and predictive analytics of manufacturing machines \cite{zhao2019deep} still remain to be explored. The recurrent neural network modules in Siamese architecture can form a good solution to analyze time-series data. This can be used to identify the similarity and differences between two instances. IIoT and machine learning, in general, can favor the use of various types of raw sensor data to allow intelligent decision making in real-time in modern industries. A large amount of this data is unstructured and the few-shot machine learning approaches have the potential to effectively use this data to get valuable insights.

\bibliographystyle{elsarticle-harv}
\bibliography{bibliography_db}

\begin{thebibliography}{47}
\expandafter\ifx\csname natexlab\endcsname\relax\def\natexlab#1{#1}\fi
\providecommand{\url}[1]{\texttt{#1}}
\providecommand{\href}[2]{#2}
\providecommand{\path}[1]{#1}
\providecommand{\DOIprefix}{doi:}
\providecommand{\ArXivprefix}{arXiv:}
\providecommand{\URLprefix}{URL: }
\providecommand{\Pubmedprefix}{pmid:}
\providecommand{\doi}[1]{\href{http://dx.doi.org/#1}{\path{#1}}}
\providecommand{\Pubmed}[1]{\href{pmid:#1}{\path{#1}}}
\providecommand{\bibinfo}[2]{#2}
\ifx\xfnm\relax \def\xfnm[#1]{\unskip,\space#1}\fi
%Type = Article
\bibitem[{Abdel-Qader et~al.(2003)Abdel-Qader, Abudayyeh and
  Kelly}]{abdel2003analysis}
\bibinfo{author}{Abdel-Qader, I.}, \bibinfo{author}{Abudayyeh, O.},
  \bibinfo{author}{Kelly, M.E.}, \bibinfo{year}{2003}.
\newblock \bibinfo{title}{{Analysis of edge-detection techniques for crack
  identification in bridges}}.
\newblock \bibinfo{journal}{Journal of Computing in Civil Engineering}
  \bibinfo{volume}{17}, \bibinfo{pages}{255--263}.
%Type = Article
\bibitem[{Altae-Tran et~al.(2017)Altae-Tran, Ramsundar, Pappu and
  Pande}]{altae2017low}
\bibinfo{author}{Altae-Tran, H.}, \bibinfo{author}{Ramsundar, B.},
  \bibinfo{author}{Pappu, A.S.}, \bibinfo{author}{Pande, V.},
  \bibinfo{year}{2017}.
\newblock \bibinfo{title}{{Low data drug discovery with one-shot learning}}.
\newblock \bibinfo{journal}{ACS central science} \bibinfo{volume}{3},
  \bibinfo{pages}{283--293}.
%Type = Article
\bibitem[{Cha et~al.(2017)Cha, Choi and
  B{\"u}y{\"u}k{\"o}zt{\"u}rk}]{cha2017deep}
\bibinfo{author}{Cha, Y.J.}, \bibinfo{author}{Choi, W.},
  \bibinfo{author}{B{\"u}y{\"u}k{\"o}zt{\"u}rk, O.}, \bibinfo{year}{2017}.
\newblock \bibinfo{title}{{Deep learning-based crack damage detection using
  convolutional neural networks}}.
\newblock \bibinfo{journal}{Computer-Aided Civil and Infrastructure
  Engineering} \bibinfo{volume}{32}, \bibinfo{pages}{361--378}.
%Type = Inproceedings
\bibitem[{Choi et~al.(2018)Choi, Jin~Chang, Fischer, Yun, Lee, Jeong, Demiris
  and Young~Choi}]{choi2018context}
\bibinfo{author}{Choi, J.}, \bibinfo{author}{Jin~Chang, H.},
  \bibinfo{author}{Fischer, T.}, \bibinfo{author}{Yun, S.},
  \bibinfo{author}{Lee, K.}, \bibinfo{author}{Jeong, J.},
  \bibinfo{author}{Demiris, Y.}, \bibinfo{author}{Young~Choi, J.},
  \bibinfo{year}{2018}.
\newblock \bibinfo{title}{{Context-aware deep feature compression for
  high-speed visual tracking}}, in: \bibinfo{booktitle}{Proceedings of the IEEE
  Conference on Computer Vision and Pattern Recognition}, pp.
  \bibinfo{pages}{479--488}.
%Type = Inproceedings
\bibitem[{Chopra et~al.(2005)Chopra, Hadsell, LeCun
  et~al.}]{chopra2005learning}
\bibinfo{author}{Chopra, S.}, \bibinfo{author}{Hadsell, R.},
  \bibinfo{author}{LeCun, Y.}, et~al., \bibinfo{year}{2005}.
\newblock \bibinfo{title}{Learning a similarity metric discriminatively, with
  application to face verification}, in: \bibinfo{booktitle}{CVPR (1)}, pp.
  \bibinfo{pages}{539--546}.
%Type = Article
\bibitem[{Devlin et~al.(2018)Devlin, Chang, Lee and Toutanova}]{devlin2018bert}
\bibinfo{author}{Devlin, J.}, \bibinfo{author}{Chang, M.W.},
  \bibinfo{author}{Lee, K.}, \bibinfo{author}{Toutanova, K.},
  \bibinfo{year}{2018}.
\newblock \bibinfo{title}{{BERT: Pre-training of deep bidirectional
  transformers for language understanding}}.
\newblock \bibinfo{journal}{arXiv preprint arXiv:1810.04805} .
%Type = Article
\bibitem[{Droghini et~al.(2019)Droghini, Squartini, Principi, Gabrielli and
  Piazza}]{droghini2019audio}
\bibinfo{author}{Droghini, D.}, \bibinfo{author}{Squartini, S.},
  \bibinfo{author}{Principi, E.}, \bibinfo{author}{Gabrielli, L.},
  \bibinfo{author}{Piazza, F.}, \bibinfo{year}{2019}.
\newblock \bibinfo{title}{{Audio Metric Learning by Using Siamese Autoencoders
  for One-Shot Human Fall Detection}}.
\newblock \bibinfo{journal}{IEEE Transactions on Emerging Topics in
  Computational Intelligence} .
%Type = Article
\bibitem[{Ferguson et~al.(2018)Ferguson, Ronay, Lee and
  Law}]{ferguson2018detection}
\bibinfo{author}{Ferguson, M.K.}, \bibinfo{author}{Ronay, A.},
  \bibinfo{author}{Lee, Y.T.T.}, \bibinfo{author}{Law, K.H.},
  \bibinfo{year}{2018}.
\newblock \bibinfo{title}{{Detection and Segmentation of Manufacturing Defects
  with Convolutional Neural Networks and Transfer Learning}}.
\newblock \bibinfo{journal}{Smart and sustainable manufacturing systems}
  \bibinfo{volume}{2}.
%Type = Inproceedings
\bibitem[{Hadsell et~al.(2006)Hadsell, Chopra and
  LeCun}]{hadsell2006dimensionality}
\bibinfo{author}{Hadsell, R.}, \bibinfo{author}{Chopra, S.},
  \bibinfo{author}{LeCun, Y.}, \bibinfo{year}{2006}.
\newblock \bibinfo{title}{{Dimensionality reduction by learning an invariant
  mapping}}, in: \bibinfo{booktitle}{2006 IEEE Computer Society Conference on
  Computer Vision and Pattern Recognition (CVPR'06)},
  \bibinfo{organization}{IEEE}. pp. \bibinfo{pages}{1735--1742}.
%Type = Inproceedings
\bibitem[{He et~al.(2017)He, Gkioxari, Doll{\'a}r and Girshick}]{he2017mask}
\bibinfo{author}{He, K.}, \bibinfo{author}{Gkioxari, G.},
  \bibinfo{author}{Doll{\'a}r, P.}, \bibinfo{author}{Girshick, R.},
  \bibinfo{year}{2017}.
\newblock \bibinfo{title}{{Mask R-CNN}}, in: \bibinfo{booktitle}{Proceedings of
  the IEEE international conference on computer vision}, pp.
  \bibinfo{pages}{2961--2969}.
%Type = Inproceedings
\bibitem[{He et~al.(2016)He, Zhang, Ren and Sun}]{he2016deep}
\bibinfo{author}{He, K.}, \bibinfo{author}{Zhang, X.}, \bibinfo{author}{Ren,
  S.}, \bibinfo{author}{Sun, J.}, \bibinfo{year}{2016}.
\newblock \bibinfo{title}{{Deep residual learning for image recognition}}, in:
  \bibinfo{booktitle}{Proceedings of the IEEE conference on computer vision and
  pattern recognition}, pp. \bibinfo{pages}{770--778}.
%Type = Article
\bibitem[{He et~al.(2019a)He, Song, Dong and Yan}]{he2019semi}
\bibinfo{author}{He, Y.}, \bibinfo{author}{Song, K.}, \bibinfo{author}{Dong,
  H.}, \bibinfo{author}{Yan, Y.}, \bibinfo{year}{2019}a.
\newblock \bibinfo{title}{Semi-supervised defect classification of steel
  surface based on multi-training and generative adversarial network}.
\newblock \bibinfo{journal}{Optics and Lasers in Engineering}
  \bibinfo{volume}{122}, \bibinfo{pages}{294--302}.
%Type = Article
\bibitem[{He et~al.(2019b)He, Song, Meng and Yan}]{he2019end}
\bibinfo{author}{He, Y.}, \bibinfo{author}{Song, K.}, \bibinfo{author}{Meng,
  Q.}, \bibinfo{author}{Yan, Y.}, \bibinfo{year}{2019}b.
\newblock \bibinfo{title}{An end-to-end steel surface defect detection approach
  via fusing multiple hierarchical features}.
\newblock \bibinfo{journal}{IEEE Transactions on Instrumentation and
  Measurement} .
%Type = Article
\bibitem[{Huang and Pan(2015)}]{huang2015automated}
\bibinfo{author}{Huang, S.H.}, \bibinfo{author}{Pan, Y.C.},
  \bibinfo{year}{2015}.
\newblock \bibinfo{title}{{Automated visual inspection in the semiconductor
  industry: A survey}}.
\newblock \bibinfo{journal}{Computers in industry} \bibinfo{volume}{66},
  \bibinfo{pages}{1--10}.
%Type = Inproceedings
\bibitem[{Huang et~al.(2018)Huang, Qiu, Guo, Wang and Yuan}]{huang2018surface}
\bibinfo{author}{Huang, Y.}, \bibinfo{author}{Qiu, C.}, \bibinfo{author}{Guo,
  Y.}, \bibinfo{author}{Wang, X.}, \bibinfo{author}{Yuan, K.},
  \bibinfo{year}{2018}.
\newblock \bibinfo{title}{Surface defect saliency of magnetic tile}, in:
  \bibinfo{booktitle}{2018 IEEE 14th International Conference on Automation
  Science and Engineering (CASE)}, \bibinfo{organization}{IEEE}. pp.
  \bibinfo{pages}{612--617}.
%Type = Article
\bibitem[{Kingma and Ba(2014)}]{kingma2014adam}
\bibinfo{author}{Kingma, D.P.}, \bibinfo{author}{Ba, J.}, \bibinfo{year}{2014}.
\newblock \bibinfo{title}{Adam: A method for stochastic optimization}.
\newblock \bibinfo{journal}{arXiv preprint arXiv:1412.6980} .
%Type = Article
\bibitem[{Kiran et~al.(2018)Kiran, Thomas and Parakkal}]{kiran2018overview}
\bibinfo{author}{Kiran, B.}, \bibinfo{author}{Thomas, D.},
  \bibinfo{author}{Parakkal, R.}, \bibinfo{year}{2018}.
\newblock \bibinfo{title}{{An overview of deep learning based methods for
  unsupervised and semi-supervised anomaly detection in videos}}.
\newblock \bibinfo{journal}{Journal of Imaging} \bibinfo{volume}{4},
  \bibinfo{pages}{36}.
%Type = Inproceedings
\bibitem[{Koch et~al.(2015)Koch, Zemel and Salakhutdinov}]{koch2015siamese}
\bibinfo{author}{Koch, G.}, \bibinfo{author}{Zemel, R.},
  \bibinfo{author}{Salakhutdinov, R.}, \bibinfo{year}{2015}.
\newblock \bibinfo{title}{Siamese neural networks for one-shot image
  recognition}, in: \bibinfo{booktitle}{ICML deep learning workshop}.
%Type = Inproceedings
\bibitem[{Krizhevsky et~al.(2012)Krizhevsky, Sutskever and
  Hinton}]{krizhevsky2012imagenet}
\bibinfo{author}{Krizhevsky, A.}, \bibinfo{author}{Sutskever, I.},
  \bibinfo{author}{Hinton, G.E.}, \bibinfo{year}{2012}.
\newblock \bibinfo{title}{{ImageNet Classification with Deep Convolutional
  Neural Networks}}, in: \bibinfo{booktitle}{Advances in neural information
  processing systems}, pp. \bibinfo{pages}{1097--1105}.
%Type = Article
\bibitem[{Lee(1991)}]{lee1991handwritten}
\bibinfo{author}{Lee, Y.}, \bibinfo{year}{1991}.
\newblock \bibinfo{title}{Handwritten digit recognition using k
  nearest-neighbor, radial-basis function, and backpropagation neural
  networks}.
\newblock \bibinfo{journal}{Neural computation} \bibinfo{volume}{3},
  \bibinfo{pages}{440--449}.
%Type = Article
\bibitem[{Levine et~al.(2016)Levine, Finn, Darrell and Abbeel}]{levine2016end}
\bibinfo{author}{Levine, S.}, \bibinfo{author}{Finn, C.},
  \bibinfo{author}{Darrell, T.}, \bibinfo{author}{Abbeel, P.},
  \bibinfo{year}{2016}.
\newblock \bibinfo{title}{{End-to-end training of deep visuomotor policies}}.
\newblock \bibinfo{journal}{The Journal of Machine Learning Research}
  \bibinfo{volume}{17}, \bibinfo{pages}{1334--1373}.
%Type = Inproceedings
\bibitem[{Lin et~al.(2018)Lin, Chiang and Hsu}]{lin2018capacitor}
\bibinfo{author}{Lin, Y.L.}, \bibinfo{author}{Chiang, Y.M.},
  \bibinfo{author}{Hsu, H.C.}, \bibinfo{year}{2018}.
\newblock \bibinfo{title}{Capacitor detection in pcb using yolo algorithm}, in:
  \bibinfo{booktitle}{2018 International Conference on System Science and
  Engineering (ICSSE)}, \bibinfo{organization}{IEEE}. pp.
  \bibinfo{pages}{1--4}.
%Type = Article
\bibitem[{Meng et~al.(2018)Meng, Yang, Chung, Lee and Shao}]{meng2018enhancing}
\bibinfo{author}{Meng, Y.}, \bibinfo{author}{Yang, Y.}, \bibinfo{author}{Chung,
  H.}, \bibinfo{author}{Lee, P.H.}, \bibinfo{author}{Shao, C.},
  \bibinfo{year}{2018}.
\newblock \bibinfo{title}{Enhancing sustainability and energy efficiency in
  smart factories: A review}.
\newblock \bibinfo{journal}{Sustainability} \bibinfo{volume}{10},
  \bibinfo{pages}{4779}.
%Type = Article
\bibitem[{Mery et~al.(2015)Mery, Riffo, Zscherpel, Mondrag{\'o}n, Lillo,
  Zuccar, Lobel and Carrasco}]{mery2015gdxray}
\bibinfo{author}{Mery, D.}, \bibinfo{author}{Riffo, V.},
  \bibinfo{author}{Zscherpel, U.}, \bibinfo{author}{Mondrag{\'o}n, G.},
  \bibinfo{author}{Lillo, I.}, \bibinfo{author}{Zuccar, I.},
  \bibinfo{author}{Lobel, H.}, \bibinfo{author}{Carrasco, M.},
  \bibinfo{year}{2015}.
\newblock \bibinfo{title}{{GDXray: The database of X-ray images for
  nondestructive testing}}.
\newblock \bibinfo{journal}{Journal of Nondestructive Evaluation}
  \bibinfo{volume}{34}, \bibinfo{pages}{42}.
%Type = Inproceedings
\bibitem[{Neculoiu et~al.(2016)Neculoiu, Versteegh and
  Rotaru}]{neculoiu2016learning}
\bibinfo{author}{Neculoiu, P.}, \bibinfo{author}{Versteegh, M.},
  \bibinfo{author}{Rotaru, M.}, \bibinfo{year}{2016}.
\newblock \bibinfo{title}{{Learning text similarity with siamese recurrent
  networks}}, in: \bibinfo{booktitle}{Proceedings of the 1st Workshop on
  Representation Learning for NLP}, pp. \bibinfo{pages}{148--157}.
%Type = Inproceedings
\bibitem[{Oquab et~al.(2014)Oquab, Bottou, Laptev and
  Sivic}]{oquab2014learning}
\bibinfo{author}{Oquab, M.}, \bibinfo{author}{Bottou, L.},
  \bibinfo{author}{Laptev, I.}, \bibinfo{author}{Sivic, J.},
  \bibinfo{year}{2014}.
\newblock \bibinfo{title}{{Learning and transferring mid-level image
  representations using convolutional neural networks}}, in:
  \bibinfo{booktitle}{Proceedings of the IEEE conference on computer vision and
  pattern recognition}, pp. \bibinfo{pages}{1717--1724}.
%Type = Article
\bibitem[{Park et~al.(2016)Park, Kwon, Park and Kang}]{park2016machine}
\bibinfo{author}{Park, J.K.}, \bibinfo{author}{Kwon, B.K.},
  \bibinfo{author}{Park, J.H.}, \bibinfo{author}{Kang, D.J.},
  \bibinfo{year}{2016}.
\newblock \bibinfo{title}{{Machine learning-based imaging system for surface
  defect inspection}}.
\newblock \bibinfo{journal}{International Journal of Precision Engineering and
  Manufacturing-Green Technology} \bibinfo{volume}{3},
  \bibinfo{pages}{303--310}.
%Type = Article
\bibitem[{Pedregosa et~al.(2011)Pedregosa, Varoquaux, Gramfort, Michel,
  Thirion, Grisel, Blondel, Prettenhofer, Weiss, Dubourg, Vanderplas, Passos,
  Cournapeau, Brucher, Perrot and Duchesnay}]{scikitlearn}
\bibinfo{author}{Pedregosa, F.}, \bibinfo{author}{Varoquaux, G.},
  \bibinfo{author}{Gramfort, A.}, \bibinfo{author}{Michel, V.},
  \bibinfo{author}{Thirion, B.}, \bibinfo{author}{Grisel, O.},
  \bibinfo{author}{Blondel, M.}, \bibinfo{author}{Prettenhofer, P.},
  \bibinfo{author}{Weiss, R.}, \bibinfo{author}{Dubourg, V.},
  \bibinfo{author}{Vanderplas, J.}, \bibinfo{author}{Passos, A.},
  \bibinfo{author}{Cournapeau, D.}, \bibinfo{author}{Brucher, M.},
  \bibinfo{author}{Perrot, M.}, \bibinfo{author}{Duchesnay, E.},
  \bibinfo{year}{2011}.
\newblock \bibinfo{title}{{Scikit-learn: Machine Learning in {P}ython}}.
\newblock \bibinfo{journal}{Journal of Machine Learning Research}
  \bibinfo{volume}{12}, \bibinfo{pages}{2825--2830}.
%Type = Article
\bibitem[{Pfrommer et~al.(2018)Pfrommer, Zimmerling, Liu, K{\"a}rger, Henning
  and Beyerer}]{pfrommer2018optimisation}
\bibinfo{author}{Pfrommer, J.}, \bibinfo{author}{Zimmerling, C.},
  \bibinfo{author}{Liu, J.}, \bibinfo{author}{K{\"a}rger, L.},
  \bibinfo{author}{Henning, F.}, \bibinfo{author}{Beyerer, J.},
  \bibinfo{year}{2018}.
\newblock \bibinfo{title}{Optimisation of manufacturing process parameters
  using deep neural networks as surrogate models}.
\newblock \bibinfo{journal}{Procedia CIRP} \bibinfo{volume}{72},
  \bibinfo{pages}{426--431}.
%Type = Inproceedings
\bibitem[{Ravi and Larochelle(2017)}]{ravi2017few}
\bibinfo{author}{Ravi, S.}, \bibinfo{author}{Larochelle, H.},
  \bibinfo{year}{2017}.
\newblock \bibinfo{title}{Optimization as a model for few-shot learning}, in:
  \bibinfo{booktitle}{International Conference on Learning Representations}.
%Type = Inproceedings
\bibitem[{Redmon et~al.(2016)Redmon, Divvala, Girshick and
  Farhadi}]{redmon2016you}
\bibinfo{author}{Redmon, J.}, \bibinfo{author}{Divvala, S.},
  \bibinfo{author}{Girshick, R.}, \bibinfo{author}{Farhadi, A.},
  \bibinfo{year}{2016}.
\newblock \bibinfo{title}{{You only look once: Unified, real-time object
  detection}}, in: \bibinfo{booktitle}{Proceedings of the IEEE conference on
  computer vision and pattern recognition}, pp. \bibinfo{pages}{779--788}.
%Type = Inproceedings
\bibitem[{Ren et~al.(2015)Ren, He, Girshick and Sun}]{ren2015faster}
\bibinfo{author}{Ren, S.}, \bibinfo{author}{He, K.}, \bibinfo{author}{Girshick,
  R.}, \bibinfo{author}{Sun, J.}, \bibinfo{year}{2015}.
\newblock \bibinfo{title}{{Faster R-CNN: Towards real-time object detection
  with region proposal networks}}, in: \bibinfo{booktitle}{Advances in neural
  information processing systems}, pp. \bibinfo{pages}{91--99}.
%Type = Inproceedings
\bibitem[{Richter et~al.(2017)Richter, Streitferdt and
  Rozova}]{richter2017development}
\bibinfo{author}{Richter, J.}, \bibinfo{author}{Streitferdt, D.},
  \bibinfo{author}{Rozova, E.}, \bibinfo{year}{2017}.
\newblock \bibinfo{title}{{On the development of intelligent optical
  inspections}}, in: \bibinfo{booktitle}{2017 IEEE 7th Annual Computing and
  Communication Workshop and Conference (CCWC)}, \bibinfo{organization}{IEEE}.
  pp. \bibinfo{pages}{1--6}.
%Type = Inproceedings
\bibitem[{Romera-Paredes and Torr(2015)}]{romera2015embarrassingly}
\bibinfo{author}{Romera-Paredes, B.}, \bibinfo{author}{Torr, P.},
  \bibinfo{year}{2015}.
\newblock \bibinfo{title}{{An embarrassingly simple approach to zero-shot
  learning}}, in: \bibinfo{booktitle}{International Conference on Machine
  Learning}, pp. \bibinfo{pages}{2152--2161}.
%Type = Inproceedings
\bibitem[{Ronneberger et~al.(2015)Ronneberger, Fischer and
  Brox}]{ronneberger2015u}
\bibinfo{author}{Ronneberger, O.}, \bibinfo{author}{Fischer, P.},
  \bibinfo{author}{Brox, T.}, \bibinfo{year}{2015}.
\newblock \bibinfo{title}{{U-net: Convolutional networks for biomedical image
  segmentation}}, in: \bibinfo{booktitle}{International Conference on Medical
  image computing and computer-assisted intervention},
  \bibinfo{organization}{Springer}. pp. \bibinfo{pages}{234--241}.
%Type = Article
\bibitem[{Shaban et~al.(2017)Shaban, Bansal, Liu, Essa and
  Boots}]{shaban2017one}
\bibinfo{author}{Shaban, A.}, \bibinfo{author}{Bansal, S.},
  \bibinfo{author}{Liu, Z.}, \bibinfo{author}{Essa, I.},
  \bibinfo{author}{Boots, B.}, \bibinfo{year}{2017}.
\newblock \bibinfo{title}{{One-shot learning for semantic segmentation}}.
\newblock \bibinfo{journal}{arXiv preprint arXiv:1709.03410} .
%Type = Inproceedings
\bibitem[{Simonyan and Zisserman(2015)}]{Simonyan15}
\bibinfo{author}{Simonyan, K.}, \bibinfo{author}{Zisserman, A.},
  \bibinfo{year}{2015}.
\newblock \bibinfo{title}{{Very Deep Convolutional Networks for Large-Scale
  Image Recognition}}, in: \bibinfo{booktitle}{{International Conference on
  Learning Representations (ICLR)}}.
%Type = Inproceedings
\bibitem[{Snell et~al.(2017)Snell, Swersky and Zemel}]{snell2017prototypical}
\bibinfo{author}{Snell, J.}, \bibinfo{author}{Swersky, K.},
  \bibinfo{author}{Zemel, R.}, \bibinfo{year}{2017}.
\newblock \bibinfo{title}{{Prototypical networks for few-shot learning}}, in:
  \bibinfo{booktitle}{Advances in Neural Information Processing Systems}, pp.
  \bibinfo{pages}{4077--4087}.
%Type = Article
\bibitem[{Song and Yan(2013)}]{song2013noise}
\bibinfo{author}{Song, K.}, \bibinfo{author}{Yan, Y.}, \bibinfo{year}{2013}.
\newblock \bibinfo{title}{A noise robust method based on completed local binary
  patterns for hot-rolled steel strip surface defects}.
\newblock \bibinfo{journal}{Applied Surface Science} \bibinfo{volume}{285},
  \bibinfo{pages}{858--864}.
%Type = Inproceedings
\bibitem[{Soukup and Huber-M{\"o}rk(2014)}]{soukup2014convolutional}
\bibinfo{author}{Soukup, D.}, \bibinfo{author}{Huber-M{\"o}rk, R.},
  \bibinfo{year}{2014}.
\newblock \bibinfo{title}{{Convolutional neural networks for steel surface
  defect detection from photometric stereo images}}, in:
  \bibinfo{booktitle}{International Symposium on Visual Computing},
  \bibinfo{organization}{Springer}. pp. \bibinfo{pages}{668--677}.
%Type = Article
\bibitem[{Tao et~al.(2018)Tao, Zhang, Ma, Liu and Xu}]{tao2018automatic}
\bibinfo{author}{Tao, X.}, \bibinfo{author}{Zhang, D.}, \bibinfo{author}{Ma,
  W.}, \bibinfo{author}{Liu, X.}, \bibinfo{author}{Xu, D.},
  \bibinfo{year}{2018}.
\newblock \bibinfo{title}{{Automatic metallic surface defect detection and
  recognition with convolutional neural networks}}.
\newblock \bibinfo{journal}{Applied Sciences} \bibinfo{volume}{8},
  \bibinfo{pages}{1575}.
%Type = Article
\bibitem[{Vafeiadis et~al.(2018)Vafeiadis, Dimitriou, Ioannidis, Wotherspoon,
  Tinker and Tzovaras}]{vafeiadis2018framework}
\bibinfo{author}{Vafeiadis, T.}, \bibinfo{author}{Dimitriou, N.},
  \bibinfo{author}{Ioannidis, D.}, \bibinfo{author}{Wotherspoon, T.},
  \bibinfo{author}{Tinker, G.}, \bibinfo{author}{Tzovaras, D.},
  \bibinfo{year}{2018}.
\newblock \bibinfo{title}{{A framework for inspection of dies attachment on PCB
  utilizing machine learning techniques}}.
\newblock \bibinfo{journal}{Journal of Management Analytics}
  \bibinfo{volume}{5}, \bibinfo{pages}{81--94}.
%Type = Inproceedings
\bibitem[{Vaswani et~al.(2017)Vaswani, Shazeer, Parmar, Uszkoreit, Jones,
  Gomez, Kaiser and Polosukhin}]{vaswani2017attention}
\bibinfo{author}{Vaswani, A.}, \bibinfo{author}{Shazeer, N.},
  \bibinfo{author}{Parmar, N.}, \bibinfo{author}{Uszkoreit, J.},
  \bibinfo{author}{Jones, L.}, \bibinfo{author}{Gomez, A.N.},
  \bibinfo{author}{Kaiser, {\L}.}, \bibinfo{author}{Polosukhin, I.},
  \bibinfo{year}{2017}.
\newblock \bibinfo{title}{{Attention is all you need}}, in:
  \bibinfo{booktitle}{Advances in neural information processing systems}, pp.
  \bibinfo{pages}{5998--6008}.
%Type = Inproceedings
\bibitem[{Vinyals et~al.(2016)Vinyals, Blundell, Lillicrap, Wierstra
  et~al.}]{vinyals2016matching}
\bibinfo{author}{Vinyals, O.}, \bibinfo{author}{Blundell, C.},
  \bibinfo{author}{Lillicrap, T.}, \bibinfo{author}{Wierstra, D.}, et~al.,
  \bibinfo{year}{2016}.
\newblock \bibinfo{title}{Matching networks for one shot learning}, in:
  \bibinfo{booktitle}{Advances in neural information processing systems}, pp.
  \bibinfo{pages}{3630--3638}.
%Type = Inproceedings
\bibitem[{Yosinski et~al.(2014)Yosinski, Clune, Bengio and
  Lipson}]{yosinski2014transferable}
\bibinfo{author}{Yosinski, J.}, \bibinfo{author}{Clune, J.},
  \bibinfo{author}{Bengio, Y.}, \bibinfo{author}{Lipson, H.},
  \bibinfo{year}{2014}.
\newblock \bibinfo{title}{{How transferable are features in deep neural
  networks?}}, in: \bibinfo{booktitle}{Advances in neural information
  processing systems}, pp. \bibinfo{pages}{3320--3328}.
%Type = Article
\bibitem[{Zhang et~al.(2018)Zhang, Geiger, Pohjalainen, Mousa, Jin and
  Schuller}]{zhang2018deep}
\bibinfo{author}{Zhang, Z.}, \bibinfo{author}{Geiger, J.},
  \bibinfo{author}{Pohjalainen, J.}, \bibinfo{author}{Mousa, A.E.D.},
  \bibinfo{author}{Jin, W.}, \bibinfo{author}{Schuller, B.},
  \bibinfo{year}{2018}.
\newblock \bibinfo{title}{{Deep learning for environmentally robust speech
  recognition: An overview of recent developments}}.
\newblock \bibinfo{journal}{ACM Transactions on Intelligent Systems and
  Technology (TIST)} \bibinfo{volume}{9}, \bibinfo{pages}{49}.
%Type = Article
\bibitem[{Zhao et~al.(2019)Zhao, Yan, Chen, Mao, Wang and Gao}]{zhao2019deep}
\bibinfo{author}{Zhao, R.}, \bibinfo{author}{Yan, R.}, \bibinfo{author}{Chen,
  Z.}, \bibinfo{author}{Mao, K.}, \bibinfo{author}{Wang, P.},
  \bibinfo{author}{Gao, R.X.}, \bibinfo{year}{2019}.
\newblock \bibinfo{title}{{Deep learning and its applications to machine health
  monitoring}}.
\newblock \bibinfo{journal}{Mechanical Systems and Signal Processing}
  \bibinfo{volume}{115}, \bibinfo{pages}{213--237}.

\end{thebibliography}

\end{document}